\documentclass[conference]{IEEEtran}  

\IEEEoverridecommandlockouts                              

\usepackage{verbatim}
\usepackage[pdftex]{graphicx}
\usepackage{epsfig}
\usepackage{amsmath,amssymb,amsfonts,bm,amscd} 
\usepackage{mathrsfs}
\usepackage{setspace}
\usepackage{fancyhdr}
\usepackage{algorithm} 
\usepackage{psfrag}
\usepackage[noadjust]{cite}
\usepackage{color}
\usepackage[usenames,dvipsnames,svgnames,table]{xcolor}
\usepackage{siunitx} 
\usepackage{tikz}
\usepackage{tikz-3dplot}
\usepackage{tkz-euclide}
\usepackage{adjustbox}
\usepackage[caption=false]{subfig}
\usepackage{flushend}

\usepackage{tkz-euclide}
\usepackage[european,siunitx]{circuitikz}
\usetikzlibrary{shapes,
  arrows,
  positioning,
  fit,
  calc,
  automata,
  intersections,
  trees,
  chains,
  decorations.pathreplacing,
  decorations.pathmorphing,
  decorations.text,
  backgrounds,
matrix}
\usepackage{pgf}
\usepackage{pgfplots, pgfplotstable}
\usepgfplotslibrary{dateplot,
                    groupplots,
                    fillbetween,
                    polar}
\pgfplotsset{compat=newest}

\graphicspath{{images/}} 

\usepackage[bookmarks=false]{hyperref}
\usepackage{algorithm}
\usepackage[noend]{algpseudocode}

\usepackage{fancyhdr}
\fancypagestyle{specialfooter}{%
\fancyhf{}
\fancyfoot[L]{Some test text}
\fancyfoot[C]{\thepage}
}

\usepackage{xcolor,colortbl}
\definecolor{grey1}{RGB}{192,192,192}
\definecolor{grey2}{RGB}{178,178,178}
\definecolor{grey3}{RGB}{150,150,150}
\definecolor{grey4}{RGB}{119,119,119}
\definecolor{grey5}{RGB}{77,77,77}
\definecolor{green1}{RGB}{112,173,71}
\definecolor{blue1}{RGB}{68,115,196}
\definecolor{red1}{RGB}{192,0,0}
\definecolor{yellow1}{RGB}{255,192,0}

\title{Image Generation for Efficient Neural Network Training in Autonomous Drone Racing
\thanks{This work is supported by Aarhus University, Department of Engineering
(28173).}
}

\author{
\IEEEauthorblockN{Th\'{e}o Morales}
\IEEEauthorblockA{\textit{DOPI} \\
\textit{Mauzé-sur-le-Mignon, France }\\
 t.morales@dopi-eye.fr}
\and
\IEEEauthorblockN{Andriy Sarabakha}
\IEEEauthorblockA{\textit{School of Mechanical and Aerospace Engineering} \\
\textit{Nanyang Technological University}\\
Singapore \\
andriy001@e.ntu.edu.sg}
\and
\IEEEauthorblockN{Erdal Kayacan}
\IEEEauthorblockA{\textit{Department of Engineering} \\
\textit{Aarhus University}\\
Aarhus C, Denmark \\
erdal@eng.au.dk}
}

\begin{document}

\maketitle

\begin{abstract}



Drone racing is a recreational sport in which the goal is to pass 
through a sequence of gates in a minimum amount of time, while avoiding collisions.
In autonomous drone racing, one must accomplish this task by flying fully
autonomously in an unknown environment by relying only on computer vision
methods for detecting the target gates. Due to the challenges such as
background objects and varying lighting conditions, traditional object
detection algorithms based on colour or geometry tend to fail.
Convolutional neural networks offer impressive advances in computer vision, but
require an immense amount of data to learn. Collecting this data
is a tedious process because the drone has to be flown manually, and the data
collected can suffer from sensor failures. In this work, a
semi-synthetic dataset generation method is proposed, using a combination of
real background images and randomised 3D renders of the gates, to provide a limitless
amount of training samples that do not suffer from those drawbacks. Using the
detection results, a line-of-sight guidance algorithm is used to cross the
gates. 
In several experimental real-time tests, the proposed framework successfully
demonstrates fast and reliable detection and navigation.

\end{abstract}



\section{Introduction}

Autonomous drone racing is an exciting case study that aims to motivate more
experts to develop innovative ways of solving complex problems, which are
applicable to other domains~\cite{Moon2019ISR}. It is calling not only for
breakthroughs in autonomous systems, but also for all intelligent robotic
systems~\cite{Delmerico2019ICRA}.  What makes drone racing such an interesting
challenge for autonomous unmanned aerial vehicles (UAVs), is the cumulative
complexity of each sub-problem to be solved~\cite{Jung2018RAL}, such as object
detection~\cite{Fu2016Sensors}, non-linear control~\cite{Sarabakha2019TFS} and
path planning~\cite{Camci2019AR}.

The rapid progress in the field of artificial intelligence brought a wider use of many novel concepts into robotics community, such as fuzzy logic~\cite{Sarabakha2019ASC}, reinforcement learning~\cite{Sarabakha2016CDC} and deep learning~\cite{Zhou2019ECC}.
Successively, with the recent breakthroughs in deep learning and the
development of increasingly powerful computer chips, deep convolutional neural
networks (CNNs) became the standard approach for computer vision
applications~\cite{Ciregan2012CVPR}. They offer impressive performances,
but require humongous quantities of data to learn a general representation of
their target dataset~\cite{Sun2017ICCV}.
It has been seen that collecting a dataset for drone
racing can be tedious and time-consuming, and is rarely balanced enough to allow
for good generalization of the knowledge~\cite{RodriguezHernandez2019ICUAS}.

\begin{figure}[!t]
\centering
\includegraphics[width=1\columnwidth]{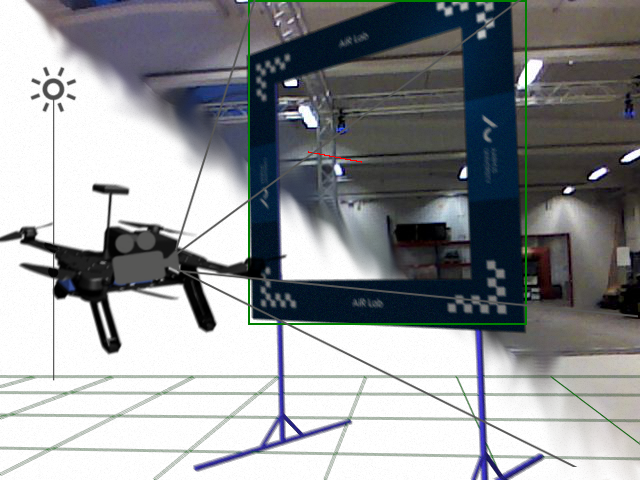}
\caption{Semi-synthetic image generated from annotated background picture. 
 A 3D render of a matched virtual scene is created with OpenGL, where
 the camera pose matches the one of the real camera used to take the
 background image. By using random picks from a dataset of base images
 annotated with their camera pose and by randomly positioning the desired
 meshes in the scene, an infinite amount of hybrid images can be produced.}
\vspace{-2mm}
\label{fig:main}
\end{figure}

Furthermore, available datasets do not
necessarily provide adequate ground truth annotations for every application,
and it can sometimes be troublesome to record a custom dataset with specific
ground truth~\cite{Jung2018AIAA}. The problem also extends to deep learning in
general, and large datasets of balanced data with specific ground truth labels
are often hard to get hands on for unconventional applications. In the case of
object detection, it might be better to create a dataset tailored to the target
application domain when it differs from common challenges. This implies that a
person has to spend a consequent amount of time annotating bounding boxes on
each object of the dataset.


In an attempt to bypass this time-consuming process, researchers have put
effort into training machine learning models directly in a simulated
environment with reinforcement learning methods~\cite{UAVRL}, or
by feeding CNNs images extracted directly from visually realistic
simulations~\cite{DDRSim}.
The common conclusion to such experiments is that the model either learns too
specifically about the synthetic domain~\cite{CNNSynthIm}, or benefits from a
performance boost when trained on a dataset of real images enriched with synthetic
ones~\cite{Rozantsev2015}. 

This work proposes a novel approach to generate a semi-synthetic dataset from a
combination of virtual scenes and real background images, to provide a limitless
amount of racing circuit combinations to train the gate detection and distance
estimation networks. The advantage of the proposed method is not limited to the
generation of random and complex placement configurations, and it also provides
automatic ground truth computation and annotation for virtually any measurement,
as well as highly realistic scenes thanks to the alliance of real-world
background photographs with modern computer graphics.
The challenges of this work lie in the credibility of the generated pictures,
where failure to render visually plausible scenes could cause the model to
overfit on the training and validation sets. In order to assess the
effectiveness of the dataset, a classical object detection CNN is trained to
detect racing gates, and a convolution-based regression network is trained on
the task of predicting the distance to the gate in meters.

This work is organised as follows. Section~\ref{sec:related} starts with a brief overview of the related works. Section~\ref{sec:dataset} describes the pipeline for the generation of semi-synthetic images. Section~\ref{sec:training} explains the approach to train the CNNs for the gate detection and distance estimation. Section~\ref{sec:setup} describes the experimental setup for real-time validation. Section~\ref{sec:detection} provides offline validation of the gate detection model and the distance estimation; while Section~\ref{sec:crossing} presents online experimental results with a quadcopter UAV for the gate crossing, to verify the robustness of the proposed approach. Finally, Section~\ref{sec:conclusion} summarizes this work with conclusions and future work.

\section{Related Work}
\label{sec:related}


An early work in~\cite{PedestrianDetection} presented a study of the use of computer
graphics to render virtual scenes and images of pedestrians, and the potential
adaptation for real-world usage of classifiers trained on such synthetic
dataset. Even though the approach is based on engineered feature detection
methods, such as histogram of gradients (HOG), and not deep convolutions, the
final statement is still valid: there is a noticeable domain shift for
classifiers trained on virtual-world data, only just as there would be with
real-world data.

Another work, in~\cite{CheungWBM17}, introduced a novel method for synthetic
pedestrian integration on unannotated real-world background images. The
novelty lies in the fact that since no annotations are used for the background
dataset, an algorithm was developed to calibrate the virtual camera, compute
the right pedestrians scale and infer a spawn probability map from features
in the image. This method allows to overlay synthetic pedestrians onto any
background image, while respecting the scale of the environment and the
obstacles in the picture. Nevertheless, this approach requires a complex pipeline of
algorithms to produce a visually coherent image.

To the best of the authors' knowledge, for the specific case of
autonomous drone racing, no occurrence of the use of synthetic datasets for the
perception logic exists. What emerged from the
literature review of the past research in this field, is that most deep
learning-based methods make use of manually collected image datasets, mostly
captured from a drone piloted by a
human~\cite{Loquercio2018, Kaufmann2019, deepracing, KimC15a}.
Unfortunately, this kind of dataset
is limited in the fact that it cannot be reused by other researchers without
having the same test environment. The same goes for the annotations and the
sensors used, which cannot be modified or enhanced without corrupting the
dataset integrity. To conclude, most researchers need to create their own dataset
when it comes to autonomous drone racing, simply because they have different
needs in terms of the required data, and that is very time-consuming.

\section{Dataset Generation}
\label{sec:dataset}

With the aim of providing a way for researchers to focus on the development of
their solution rather than the collection of a dataset, and to produce a
theoretically infinite amount of possible configurations, a
semi-synthetic dataset generation pipeline is proposed and further put to the test.
The idea proposed for the aforementioned solution is to generate a virtual
scene using OpenGL representing randomly positioned virtual gates that a
drone would potentially have to fly through in the real world. Then, a rendered image
of the scene is overlaid onto a real image, taken from the
quadcopter's camera, as to represent a random portion of a racing circuit, as
seen from the UAV's field of view. In order to do so, several requirements have
to be met.

Firstly, the virtual camera must be a perfect model of the actual camera used
to capture the background images, since the perspective calculation is derived
from its intrinsic parameters. Therefore, the drone camera needs to be
calibrated in order to estimate its intrinsic parameters, before being able to
apply them to the virtual camera.
Secondly, a motion capture system is used to record the position and
orientation of the drone in 3D space, also known as the extrinsic parameters of
the camera. This is crucial for the generation of a virtual scene whose
perspective matches perfectly the perspective of the drone so that the
generated scene can later be overlaid onto the real scene.
Finally, the dimensions of the virtual scene must be in accordance with the
actual dimensions of the physical space where the dataset of background images
was recorded, otherwise unwanted artefacts such as gates being visible outside
the confinement of the real scene could be produced. The same reasoning can be
applied to the 3D models of gates or obstacles, since their scale
matters for the usage of the generated images.

The process of generating a synthetic scene containing random meshes is done in
several steps. Firstly, a virtual camera is initialized based on the real
camera's intrinsic parameters, from which a projection matrix can be derived.
In a second time, a random choice of available meshes is spawned in the virtual
scene, with random translations and rotations. The final step is to compute
the closest gate to the camera being in its field of view, and project its
center coordinates onto the image so that a ground truth label can be inferred.
The general idea of this procedure is expressed as pseudo-code in Algorithm~\ref{alg:gen}.

\begin{algorithm}[!b]
  \caption{Semi-synthetic image generation}\label{alg:gen}
  \begin{algorithmic}[1]
    \Procedure{GenerateImage}{}
    \State $\textit{background} \gets \Call{GetRandomBaseImage}{ }$
    \State $\textit{cameraPose} \gets \Call{GetAnnotations}{\textit{background}}$
    \State $\textit{gatePoses} \gets [\,]$
    \State $i \gets \Call{Rand}{1, \textbf{MAX\_GATES}}$
    \For{$i > 0$}
      \Repeat
	\State $\textit{pose} \gets \Call{RandomPose}{ }$
	  \State $\textit{valid} \gets \textbf{True}$
	\ForAll{\textit{gatePose} in \textit{gatePoses}}
	  \State $\textit{distance} \gets \Call{Norm}{\textit{pose}, \textit{gatePose}}$
	  \If{$\textit{distance} < \textbf{MIN\_DISTANCE}$}
	    \State $\textit{valid} \gets \textbf{False}$
	  \EndIf
	\EndFor
	\Until{$\textit{valid}$}
	\State $\textit{gatePoses}[i] \gets \textit{pose}$
      \State $i \gets i-1$.
    \EndFor
    \State $\textit{scene} \gets \Call{Render}{\textit{cameraPose, gatePoses}}$
    \State $\textit{scene} \gets \Call{AddNoise}{\textit{scene}}$
    \State $\textit{scene} \gets \Call{AddMotionBlur}{\textit{scene}}$
    \State $\textit{image} \gets \textit{background} + \textit{scene}$
    \State $\textit{annotations} \gets \Call{Annotate}{gatePoses}$

    \Return $\textit{image}, \textit{annotations}$
    \EndProcedure
  \end{algorithmic}
\end{algorithm}

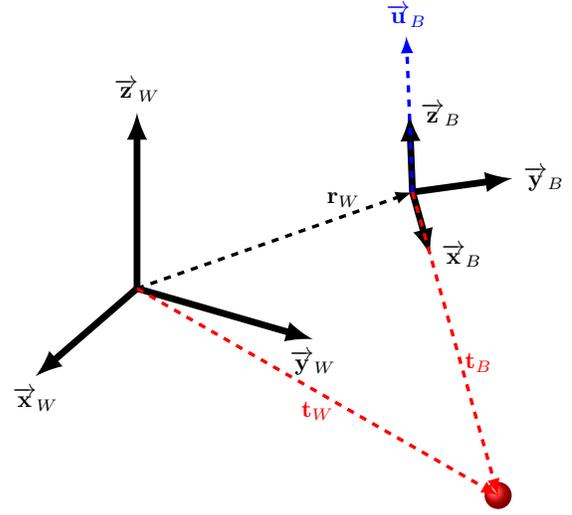
\begin{figure}[!t]
  \hspace{-1cm}
  \scalebox{0.9}{
    \pgfmathsetmacro{\viewtheta}{60}
\pgfmathsetmacro{\viewphi}{120}
\pgfmathsetmacro{\rvec}{5}
\pgfmathsetmacro{\thetavec}{60}
\pgfmathsetmacro{\phivec}{100}
\pgfmathsetmacro{\rvectarget}{5}
\pgfmathsetmacro{\thetavectarget}{90}
\pgfmathsetmacro{\phivectarget}{40}

\tdplotsetmaincoords{\viewtheta}{\viewphi}

\begin{tikzpicture}[scale=1,tdplot_main_coords,>=latex]

\tdplotsetcoord{O}{0}{0}{0}
\tdplotsetthetaplanecoords{0}
\draw[->,line width=1mm] (0,0,0) -- (3,0,0) node[anchor=north]{$\overrightarrow{\mathbf{x}}_{W}$};
\draw[->,line width=1mm] (0,0,0) -- (0,3,0) node[anchor=north]{$\overrightarrow{\mathbf{y}}_{W}$};
\draw[->,line width=1mm] (0,0,0) -- (0,0,3) node[anchor=south]{$\overrightarrow{\mathbf{z}}_{W}$};

\tdplotsetcoord{P}{\rvec}{\thetavec}{\phivec}
\draw[dashed,->,line width=0.5mm] (O) -- (P) node[near end,anchor=south]{$\mathbf{r}_{W}$};

\tdplotsetrotatedcoordsorigin{(P)}
\tdplotsetrotatedcoords{20}{10}{20}
\draw[tdplot_rotated_coords,->,line width=1mm] (0,0,0) -- (1.5,0,0) node[anchor=west]{$\overrightarrow{\mathbf{x}}_{B}$};
\draw[tdplot_rotated_coords,->,line width=1mm] (0,0,0) -- (0,1.5,0) node[anchor=west]{$\overrightarrow{\mathbf{y}}_{B}$};
\draw[tdplot_rotated_coords,->,line width=1mm] (0,0,0) -- (0,0,1.5) node[anchor=west]{$\overrightarrow{\mathbf{z}}_{B}$};
\draw[blue,dashed,tdplot_rotated_coords,->,line width=0.5mm] (0,0,0) -- (0,0,3) node[anchor=south]{$\overrightarrow{\mathbf{u}}_{B}$};

\tdplotsetcoord{B}{11}{80}{59.5}
\tdplotsetrotatedcoordsorigin{(B)}
\shade[ball color=red,tdplot_rotated_coords] (0,0,0) circle (0.2cm);

\draw[red,dashed,->,line width=0.5mm] (P) -- (B)
  node[anchor=north west,midway]{$\mathbf{t}_{B}$};

\draw[red,dashed,->,line width=0.5mm] (O) -- (B) node[anchor=north,midway]{$\mathbf{t}_{W}$};

\end{tikzpicture}
  }
  \caption{World frame and body frame camera coordinates. A target vector is computed from the camera extrinsics for the OpenGL camera initialization.}
  \label{fig:coordinatesystem}
\end{figure}

In an OpenGL scene, the virtual camera is initialized with an up vector, an eye
vector, and a target vector, as illustrated in Fig.~\ref{fig:coordinatesystem}. Those are passed to a helper function that
computes the view matrix responsible for the 3D transformations.
The target is a point in space where
the camera is looking at; it is calculated by the Hamilton product of the drone
orientation quaternion ($\mathbf{q}_B^W$) with the unit vector $\mathbf{\hat{x}}_B$ on the $\mathbf{x}_B$-axis (to be in front of
the field of view) in the body frame, added to the drone translation, $\mathbf{r}_W$ in the
world frame. Its final expression is as follows:
\begin{equation} \label{equ:targetvector}
	\mathbf{t}_W = \mathbf{r}_W + \mathbf{q}_B^W\:\mathbf{\hat{x}}_B\:
	(\mathbf{q}_{B}^{W})^{-1}.
\end{equation}
As for the up vector $\mathbf{u}_W$, it is, in the body frame, a unit vector ($\mathbf{\hat{z}}_B$)
orthogonal to the camera's target vector, and calculated by the Hamilton
product of the drone orientation quaternion $\mathbf{q}_B^W$ and a unit vector $\mathbf{\hat{z}}_B$ on the $\mathbf{z}_B$-axis.
The final up vector in the world frame is expressed as follows:
\begin{equation} \label{equ:upvector}
	\mathbf{u}_W = \mathbf{r}_W + \mathbf{q}_B^W \:\mathbf{\hat{z}}_B\:
	(\mathbf{q}_B^W)^{-1}.
\end{equation}

Once the camera is modelled and is positioned in the virtual scene, so that it
matches the real world conditions, each randomly picked 3D mesh is translated
and rotated in the scene. By applying random transformations, the algorithm
aims to recreate unique combinations of gates for each frame.
As a result, it is possible, and often the case, that only a small subset of
the originally selected meshes are visible on the image frame (depending on the
camera pose), or even none of them. For instance, if an extracted frame from the
base dataset indicates that the drone is facing a wall up close, it would make
sense not to render any gates at all to preserve coherence. In order to
achieve this, the virtual scene is constrained within boundaries that match the
physical environment in which the background images were recorded. The origin
of the motion capture system being roughly at the centre of the premise, it is
trivial to set boundaries around the origin of the virtual world, that match
the ones of the physical environment.

\begin{figure*}[!t]
\centering
\subfloat[Background only.]{\includegraphics[width=0.65\columnwidth]{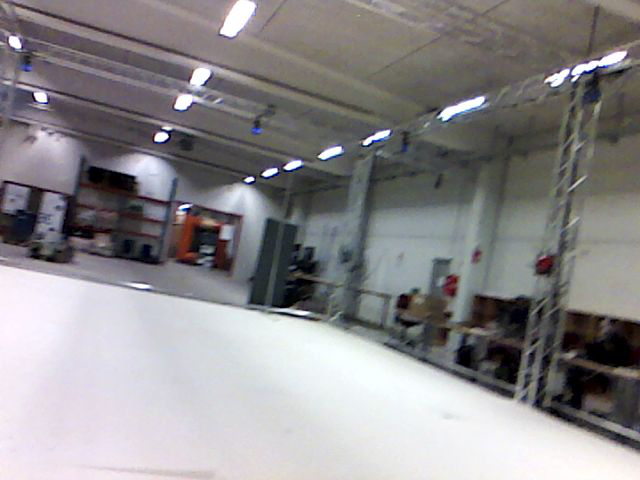}%
\label{fig:bg}} \hfil
\subfloat[Projection only.]{\includegraphics[width=0.65\columnwidth]{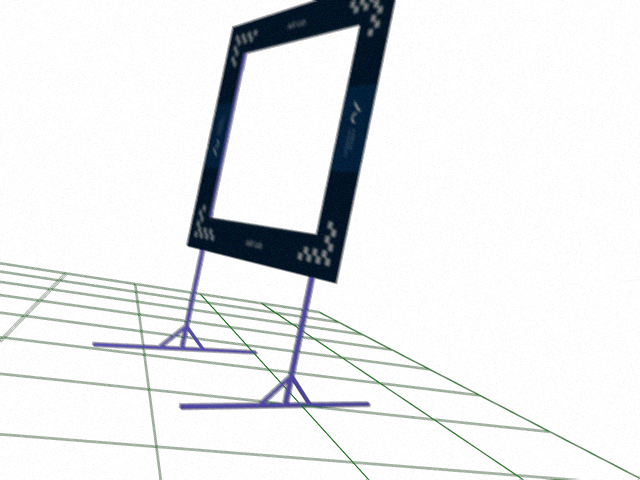}%
\label{fig:pj}} \hfil
\subfloat[Combined image.]{\includegraphics[width=0.65\columnwidth]{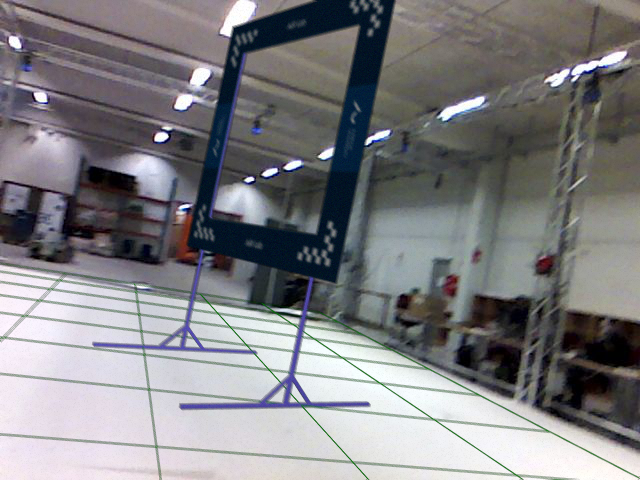}%
\label{fig:cb}}
\caption{The steps of a semi-synthetic image generation. Decomposition
	of the making of a semi-synthetic image, representing randomly positioned
	3D models of randomly selected gates overlaid on top of a randomly selected
	real-world background image. The virtual scene's camera is positioned and
	oriented using the exact configuration of the background image so that
	both perspectives match. The perspective grid is kept for visualization purposes.}
\label{fig:pipeline}
\end{figure*}

Once the virtual scene is ready to be rendered, the annotations can be
computed, namely: every gate's bounding box coordinates and
their distance to the camera. In order to compute the bounding box coordinates
in the image frame, the gate centre has to be determined in the world frame,
by simply applying the random 3D transformations to the centre vector parsed
from a configuration file, for each mesh. From that same file, the width and
height of each mesh are provided, allowing to compute easily the corners of the
bounding box in the world frame.
Lastly, the image frame coordinates (in pixel) can be
obtained by applying the viewport transform:
\begin{equation}
	\begin{bmatrix}
		x_w\\y_w
	\end{bmatrix}
	=
	\begin{bmatrix}
		\cfrac{w}{2} x_n + x + \cfrac{w}{2}\\
		\cfrac{h}{2} y_n + y + \cfrac{h}{2}\\
	\end{bmatrix},
\end{equation}
where $x_w$ and $y_w$ are the gate centre coordinates in the window
frame (or image) and $x_n$ and $y_n$ in the normalised device
coordinate system, $w$ and $h$ are the width and height of the view port -- or
the output image frame -- and $x$ and $y$ correspond to the viewport offset in
pixels, which are initially set to 0.
Furthermore, to improve the correctness of the annotations and follow a human
logic, an out-of-screen tolerance is added to select a target gate
even if it is slightly outside of the viewport, by requiring that at least three
corners of the gate be inside the frame. In that way, the intuition of
targeting a gate which is clearly visible, but whose centre does not seem to be
in the field of view, can be applied to the network.

The final step of forging semi-synthetic images is to merge the virtual scene
into a real environment, and to blend the two images in a way that it looks as if it
were real, as depicted in Fig.~\ref{fig:pipeline}. First and foremost, a number of image
deformations and distortions are applied to the rendered virtual scene. This is
needed because real cameras cannot capture reality as it can be done in a
controlled and sterile environment that can provide computer graphics. In
particular, the camera used on the UAV produces noisy images, has a rather
low resolution, does not perform well in high dynamic range and is highly
subject to the motion blur. If the generated image of a still and sharp gate is
simply overlaid on top of a blurry and noisy background, it will produce an
incoherent and unnatural result. In order to apply a simulated motion blur, the
amount of motion blur present in the background image is first estimated. This
is done via the Laplacian operator, or the second order partial derivative
defined as follows:
\begin{equation} \label{equ:laplacianoperator}
    \nabla f = \frac{\partial^2 f}{\partial x^2}
        + \frac{\partial^2 f}{\partial y^2},
\end{equation}
where $f$ is the pixel intensity. This operator detects high changes in the pixel intensity, or in other words: edges.
A low amount of detected edges in the image reflects in a blurry image, where the
pixel intensity is more evenly distributed. That is why the variance of the
Laplacian of the image is used with a set of three thresholds, found by trial
and error, to apply a different blur filter accordingly. Each of the three blur
kernels has different coefficients, meaning that a different amount of
synthetic motion blur is applied to the generated image by convolution,
depending on the background image. Lastly, a reasonable amount of
Gaussian-distributed additive noise is incorporated to the generated image. The mean value is set to $0$, while the variance is found by trial and error.

\section{Convolutional Neural Networks Training}
\label{sec:training}



\begin{figure}[!b]
  \centering
  \subfloat[Base image from the camera with bounding box annotation.]{
    \includegraphics[width=0.8\columnwidth]{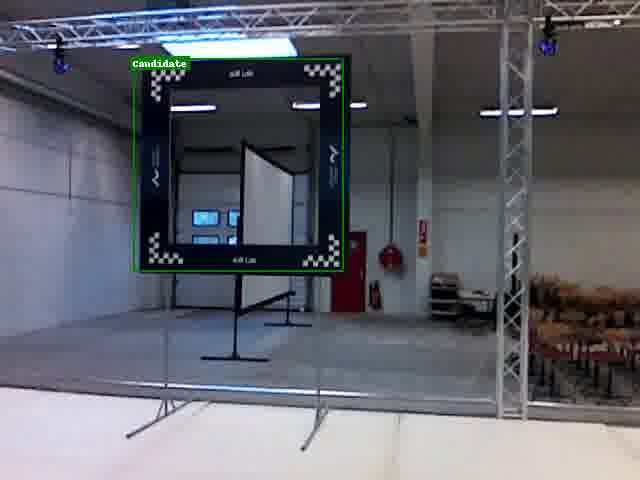}
    \label{fig:cropping1}
  } \hfil
  \subfloat[Cropped image after pre-processing.]{
    \includegraphics[width=0.8\columnwidth]{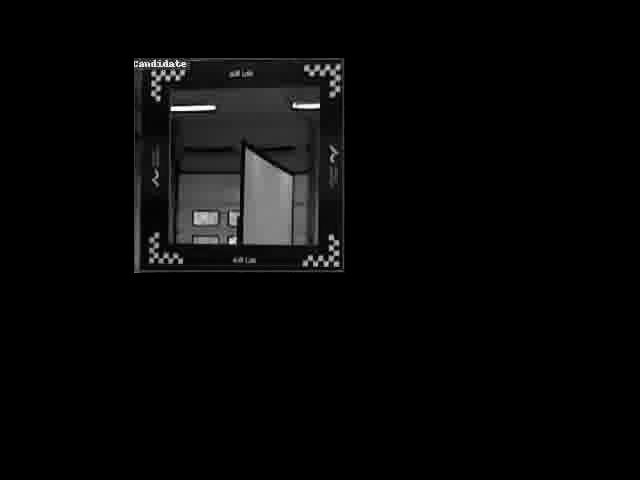}
    \label{fig:cropping2}
  }
  \caption{Region of interest extraction by cropping. The
  resulting image holds the essence of the distance information.}
  \label{fig:cropping}
\end{figure}

A single-shot detector is trained on the
semi-synthetic dataset previously generated in Section~\ref{sec:dataset}, and predicts bounding box
coordinates for the closest visible gate. The given bounding box encloses the
entire gate without its legs, as shown in Fig.~\ref{fig:cropping1}, and therefore the centre of this region
corresponds to the desired crossing point.

The network used for this task is the light version of SSD~\cite{Liu2016ECCV}, called
SSDLite, with MobileNetV2~\cite{Sandler2018CVPR} as the backbone
feature extraction network. This choice is justified by the relatively
low computation needs for a reasonably high detection precision. To train the
network for this specific object detection task, transfer learning is used
and the network architecture is left unchanged. The process only requires to change the input image resolution, the
number of classes, and the different aspect ratios per layer (used for
size and rotation invariance) which were set to globally match the aspect ratio
of the gates at different orientations and distances to the camera. 

A total of
three classes were used to classify the gates: target, facing front and
facing back.
During training, the model's performance is monitored using a semi-synthetic
validation set, using $40\text{K}$ training samples for $4\text{K}$ validation samples.
The domain shift is observed when evaluating the model on the test dataset,
consisting of around $17\text{K}$ annotated images of real gates.

\begin{figure}[!b]
    \centering
    \includegraphics[width=0.99999\columnwidth]{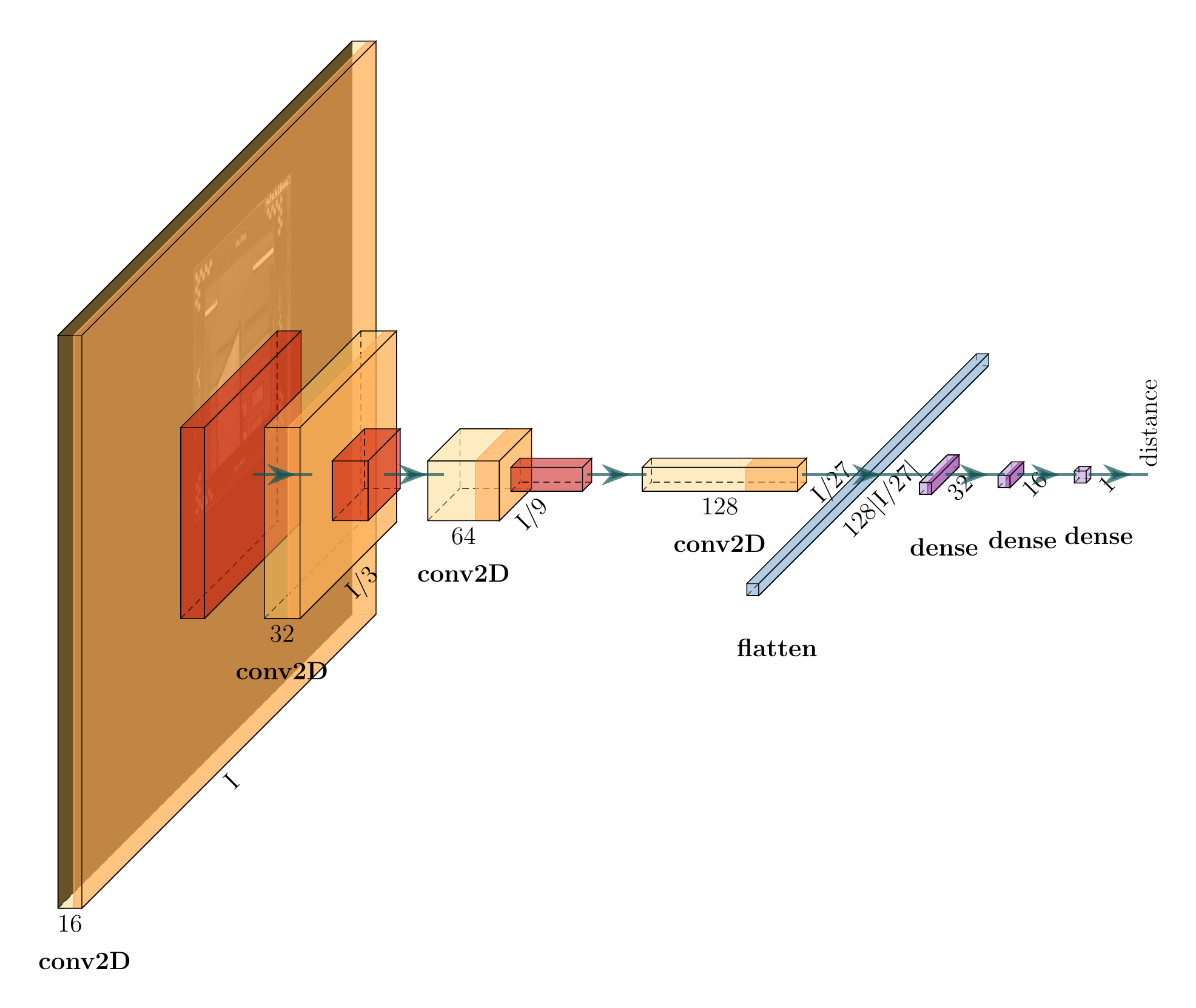}
    \caption{Topology of the proposed CNN for the estimation of the distance to the target gate. The network consists of four convolution layers with max pooling and ReLU activation functions, one flatten layer and three fully-connected layers with ReLU activation functions.}
    \label{fig:distance_cnn}
\end{figure}

Furthermore, a second CNN, presented in Fig.~\ref{fig:distance_cnn}, is needed to estimate the
distance to the target gate. Indeed, a critical issue arises when the
quadcopter gets closer to the gate: it becomes invisible in the camera's FOV
and the CNN can no longer detect it. To remedy the problem, the distance to
the target gate is periodically checked during the flight, and when short
enough, the crossing stage is initiated. This network is trained on a slightly
different synthetic dataset consisting of grey scale images of a single gate in
random orientations and distances.

A pre-processing step is cropping the
bounding box of the gate as to replace the
rest of the image with black pixels, as shown in Fig.~\ref{fig:cropping2}. That
way, only the relevant information is fed into the network, and the output is a
floating-point value corresponding to the distance in meters.
Finally, the perception module is implemented as a ROS node which runs the two
neural networks, and does not apply any filtering to the bounding box
predictions. The information published by the node contains the bounding box
coordinates along with the estimated distance to the target.

\section{Experimental Setup}
\label{sec:setup}

To validate the capabilities of the proposed methodology in
Section~\ref{sec:detection}, the Intel\textsuperscript{\textregistered} Aero
Ready-to-Fly drone was chosen due to its compact
size and flight characteristics that are suitable for a drone racing
application. This UAV is geared for developers and researchers who desire a
fast path to getting applications airborne. It is equipped with the
Intel\textsuperscript{\textregistered} Aero Compute Board (CPU:
Intel\textsuperscript{\textregistered} Atom\textsuperscript{\texttrademark}
x7-Z8750; $4 \si{GB}$ LPDDR3-1600, $32 \si{GB}$ eMMC) where Linux  was
installed; while the robot operating system, ROS Kinetic, is used to
communicate with the UAV. Moreover, the Vicon motion capture system is used to
compute the UAV's real-time velocity. The latter
is fed into the extended Kalman filter (EKF) running on the Intel Aero Flight
Controller with a Dronecode PX4 autopilot to obtain more accurate velocity
information through sensor fusion with the data coming from inertial measurement unit.
In addition, the Intel\textsuperscript{\textregistered}
RealSense\textsuperscript{\texttrademark} R200 camera is used to provide an RGB
video stream. This information is fed into the ground station computer (CPU
(i5): $2.7 \si{GHz}$, $64 \si{bit}$, quad-core; GPU (GeForce 940MX): $1
\si{GHz}$, $1 \si{GB}$ GDDR5; RAM: $7 \si{GB}$ DDR4) where the gate detection
and distance estimation algorithms are executed.
Moreover, the LeddarOne range finder is used to detect the gate
crossing instant.

As for the gates to be crossed, they are made of laser-printed cardboard
material glued onto a metal frame of size $1.5\si{m} \times 1.5\si{m}$. To
provide ground truth annotations for the experiments, the motion capture system
is used to record the poses of the gates and the UAV. The 3D models used for
the generation of the synthetic images are $1:1$ polygon mesh representations of the
actual gates.

\section{Gate Detection}
\label{sec:detection}

The performance of the gate detection model is evaluated on two datasets: a
first semi-synthetic test dataset of 17K images and a test dataset consisting
of 17K images of real gates. The latter was recorded with the UAV's camera, and
using the annotated 3D poses provided by the motion capture system, ground
truth bounding boxes were automatically computed for the target gate only.
Table~\ref{tab:detection} shows the {average precision} (AP) and
{average recall} (AR) values for both datasets, with different
{intersection over union} (IoU) threshold values, and for the same class
representing the closest gate in the image. 

\begin{table}[!b]
  \centering
  \caption{Average precision and recall for different IoU
  thresholds.}
  \begin{tabular}{r|l|l|l|l}
      IoU threshold & \multicolumn{2}{c|}{Semi-synthetic} & \multicolumn{2}{c}{Real} \\
                    & AP & AR & AP & AR \\
      \hline
      0.50 & 0.891 & 0.647 & \textbf{0.905} & 0.576\\
      0.75 & \textbf{0.797} & 0.601 & 0.648 & 0.471\\
      0.90 & \textbf{0.436} & 0.396 & 0.011 & 0.050\\
  \end{tabular}
  \label{tab:detection}
\end{table}

A higher IoU requires the predicted
bounding box to closer match the ground truth, while a lower value allows for a
larger error margin.  The results in Table~\ref{tab:detection} show an overall
higher detection precision for the synthetic dataset, which is expected
considering that the network was trained on such data. However, it is important
to denote that the ground truth annotations for the test dataset are
biased. In fact, several factors are responsible for the
shift in the ground truth bounding boxes, namely: inaccurate gate 3D pose
caused by manually recording it with a hand-held tracker, occluded UAV during
the manual flight, network lag causing packet loss, and other possible causes.
Nevertheless, an average precision of $0.9$ for an IoU of $0.5$ is a satisfying
result proving that such model can be used for real time detection.
Finally, Table~\ref{tab:detection-detailed} shows the mean average precision
(mAP), AP and AR for the three classes evaluated on the synthetic test set
(only the target gate ground truth could be computed for the real set). It
shows that the model is well-balanced, which reflects the dataset it
was trained on. This is rarely the case when manually collecting a dataset, and
often leads to a biased predictor.

\begin{table}[!b]
  \centering
  \caption{Average precision for different IoU
    thresholds and for every class. (synthetic dataset only)}
  \begin{tabular}{r|l|l|l}
    Class & AP$_{0.5}$ & AP$_{0.75}$ & AP$_{0.9}$\\
      \hline
      Target gate & 0.891 & 0.797 & 0.436 \\
      Forward gate & 0.819 & 0.726 & 0.231\\
      Backward gate & 0.907 & 0.811 & 0.382\\
      \hline
      mAP & 0.872 & 0.778 & 0.35\\
  \end{tabular}
  \label{tab:detection-detailed}
\end{table}

Regarding the distance estimation, the model is evaluated on the same real test
dataset and an equally large synthetic dataset that contains only images of
single gates using the {mean absolute error} (MAE) in meters and an accuracy measure computed with respect to
different error thresholds, as shown on Table~\ref{tab:distance}.
One could interpret those error measurements to be high, however, precision is
not required for this application, and $0.66 \si{m}$ of mean error proved to be
sufficient for the crossing condition of the state machine.

\begin{table}[!b]
  \centering
  \caption{Mean absolute error (in meters) and accuracy for different distance
    error thresholds.}
  \begin{tabular}{r|l|l|l|l}
    Error threshold & \multicolumn{2}{c|}{Semi-synthetic} & \multicolumn{2}{c}{Real} \\
    (in meters)     & MAE & Accuracy & MAE & Accuracy \\
      \hline
      0.75 & 0.401 & 0.854 & 0.660 & 0.660 \\
      0.50 & 0.401 & 0.722 & 0.660 & 0.505 \\
      0.25 & 0.401 & 0.445 & 0.660 & 0.283 \\
  \end{tabular}
  \label{tab:distance}
\end{table}

\section{Gate Crossing}
\label{sec:crossing}

\begin{figure}[!b]
  \centering
  \scalebox{0.75}{
    \tikzstyle{int}=[draw, align=center, fill=blue!0, minimum size=2em, text width=.75cm]
\tikzstyle{sum}=[draw, fill=blue!0, shape=circle, inner sep=0.3pt]
\begin{tikzpicture}[node distance=1cm,auto,>=latex]    
        \node [int, line width=0.5mm, text width=1.5cm, minimum height=1cm, fill=red!20] (cnn1) {Gate Detector};
        \node [int, line width=0.5mm, below = 1cm of cnn1, text width=1.5cm, minimum height=1cm, fill=red!20] (cnn2) {Distance Estimator};
        \begin{scope}[on background layer]\node [fit = (cnn1) (cnn2), dashed, draw, inner sep=0.2cm, fill=red!10] (perception) {};\end{scope}
        \node [above = 0.1cm of perception, align = center] (label1) {Perception};
        \node [int, line width=0.5mm, right = 2cm of cnn1, text width=1.5cm, minimum height=1cm, fill=green!20] (pid) {Velocity Controller}; 
        \node [int, line width=0.5mm, below = 1cm of pid, text width=1.5cm, minimum height=1cm, fill=green!20] (state) {State Machine};
        \begin{scope}[on background layer]\node [fit = (pid) (state), dashed, draw, inner sep=0.2cm, fill=green!10] (navigation) {};\end{scope}
        \node [above = 0.1cm of navigation, align = center] (label2) {Navigation};
        \node [int, line width=0.5mm, right = 2cm of state, text width=1.5cm, minimum height=1cm, fill=blue!20] (controller) {Low-level Controller};       
        \node [int, line width=0.5mm, above = 0.5cm of controller, text width=1.5cm, minimum height=1cm, fill=blue!20] (lidar) {Vertical Lidar};
        \node [int, line width=0.5mm, above = 0.5cm of lidar, text width=1.5cm, minimum height=1cm, fill=blue!20] (camera) {RGB Camera};
        \begin{scope}[on background layer]\node [fit = (controller) (camera) (lidar), dashed, draw, inner sep=0.2cm, fill=blue!10] (uav) {};\end{scope}
        \node [above = 0.1cm of uav, align = center] (label3) {Drone};
                   
		\draw[->, line width=0.5mm] (cnn1.south) -- node[right, pos=0.5, align=left] {gate\\region} (cnn2.north);
		\draw[->, line width=0.5mm] (cnn1.east) -- node[above, pos=0.5, align=center] {gate\\center} (pid.west);
		\draw[->, line width=0.5mm] (cnn1.east) -| ($(state.west) + (-1cm, 0.2cm)$) -- ($(state.west) + (0, 0.2cm)$);
		\draw[->, line width=0.5mm] ($(cnn2.east) + (0, -0.2cm)$) -- node[below, pos=0.5, align=center] {gate\\distance} ($(state.west) + (0, -0.2cm)$);
		\draw[->, line width=0.5mm] ($(pid.south) + (-0.2cm, 0)$) -- node[right, pos=0.5] {velocity} ($(state.north) + (-0.2cm, 0)$);
		\draw[->, line width=0.5mm] (state.east) -- node[above, pos=0.5, align=center] {velocity\\command} (controller.west);	
		\draw[->, line width=0.5mm] (lidar.east) -| ($(lidar.east) + (0.5cm, -2.5cm)$) -| node[above, pos=0.3] {altitude} ($(state.south) + (-0.2cm, 0)$);
		\draw[->, line width=0.5mm] (camera.east) -| ($(camera.east) + (1cm, -4.5)$) -| node[above, pos=0.45] {image} ($(cnn2.west) + (-0.5cm, 0)$) -- (cnn2.west);
		\draw[->, line width=0.5mm] (camera.east) -| ($(camera.east) + (1cm, -4.5)$) -| ($(cnn1.west) + (-0.5cm, 0)$) -- (cnn1.west);

\end{tikzpicture}
  }
  \caption{Block diagram of the proposed architecture for the real-time navigation of UAV to cross the gates.}
  \label{fig:system}
\end{figure}
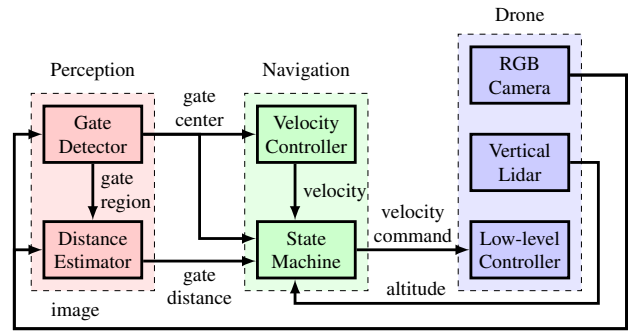

The challenge of navigating through the gates is approached as a simpler sub-problem consisting
of getting from point $A$ to point $B$, where $A$ is
the position of the last crossed gate or starting point, and $B$ is the
next gate to cross. Therefore, 
we opted for a state machine where each state represents a behaviour for a
specific stage of the flight, since it is computationally efficient
and can easily accept changes in the gate's position. As illustrated in
Fig.~\ref{fig:system}, a state machine supervises the flight and utilizes a
proportional-integral-derivative (PID) controller to generate the velocity
commands to align with the centre of the gate. This choice of approach is
motivated by the fact that admittedly, if the nearest gate detection is quick
and precise, considering a race as a succession of independent targets to be
reached, it is a robust solution that can be applied to any
environment with any gate of any size. For the tests, both gate detection and
distance estimation networks are running off-board at
$20 \si{Hz}$.


\begin{figure}[!b]
\centering
\includegraphics[width=0.99999\columnwidth]{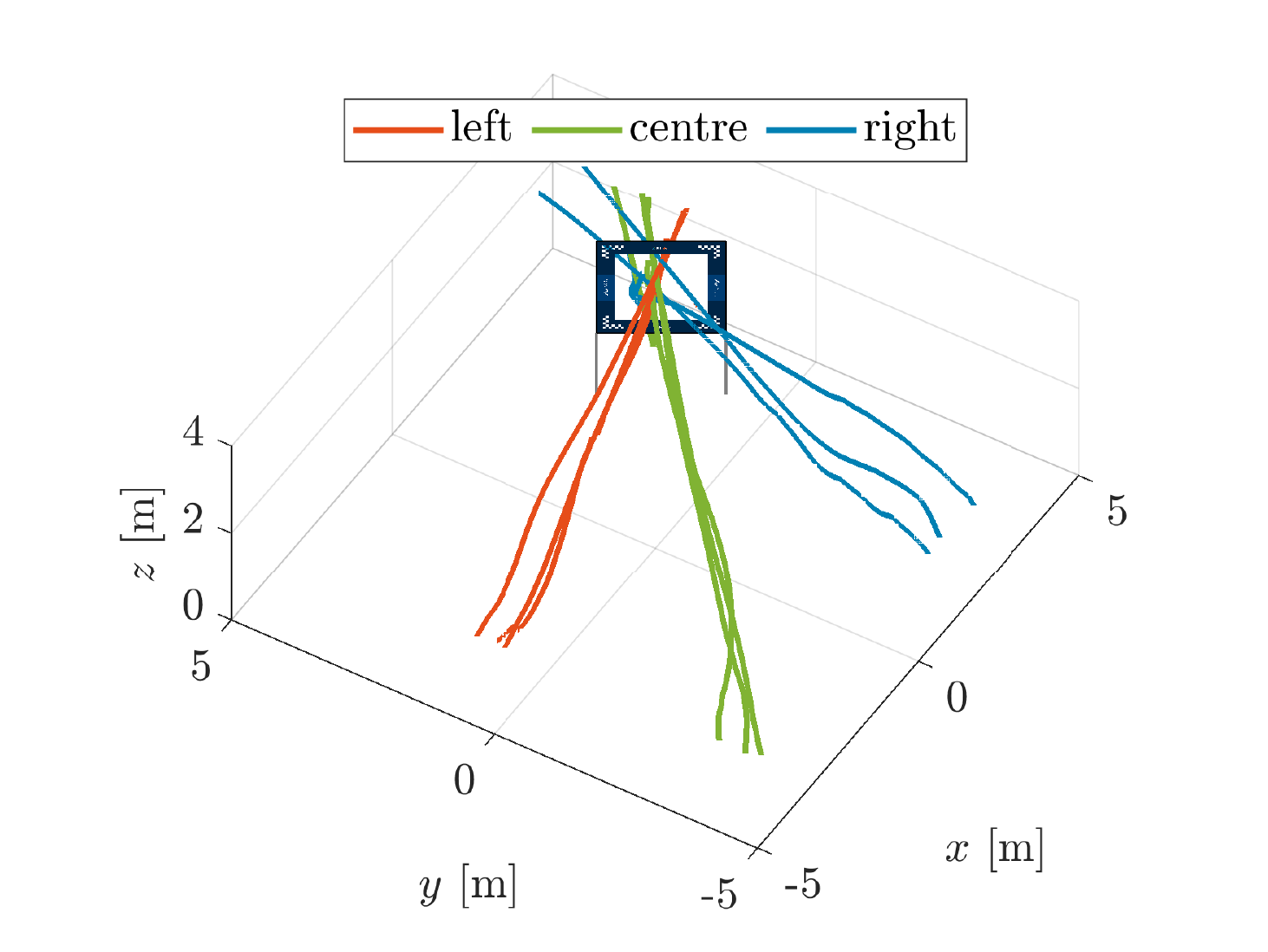}
\caption{3D view of the gate crossing at $2 \si{m/s}$ from different starting
locations. It can be observed that UAV successfully crosses the gate regardless
of its initial position.}
\label{fig:gateCrossing_3D}
\end{figure}

To prove that the proposed method can be used in drone racing, real-time
experiments are conducted, starting from three different locations (left, centre
and right with respect to the gate position) for the exercise of crossing a
single gate. Fig.~\ref{fig:gateCrossing_3D} shows the 3D view of the UAV's trajectories for different initial
locations at the flying speed of $2 \si{m/s}$. The proposed method has not been tested for higher speeds for safety
reasons, but it can be assumed by the low amount of failed attempts that the tested
speed is not exploiting the full potential of the method. It is possible to observe from Fig.~\ref{fig:gateCrossing_3D} that the proposed method is robust enough
to handle fast flights and is not limited to perpendicular trajectories.

\begin{figure}[!t]
\centering
\includegraphics[width=0.9\columnwidth]{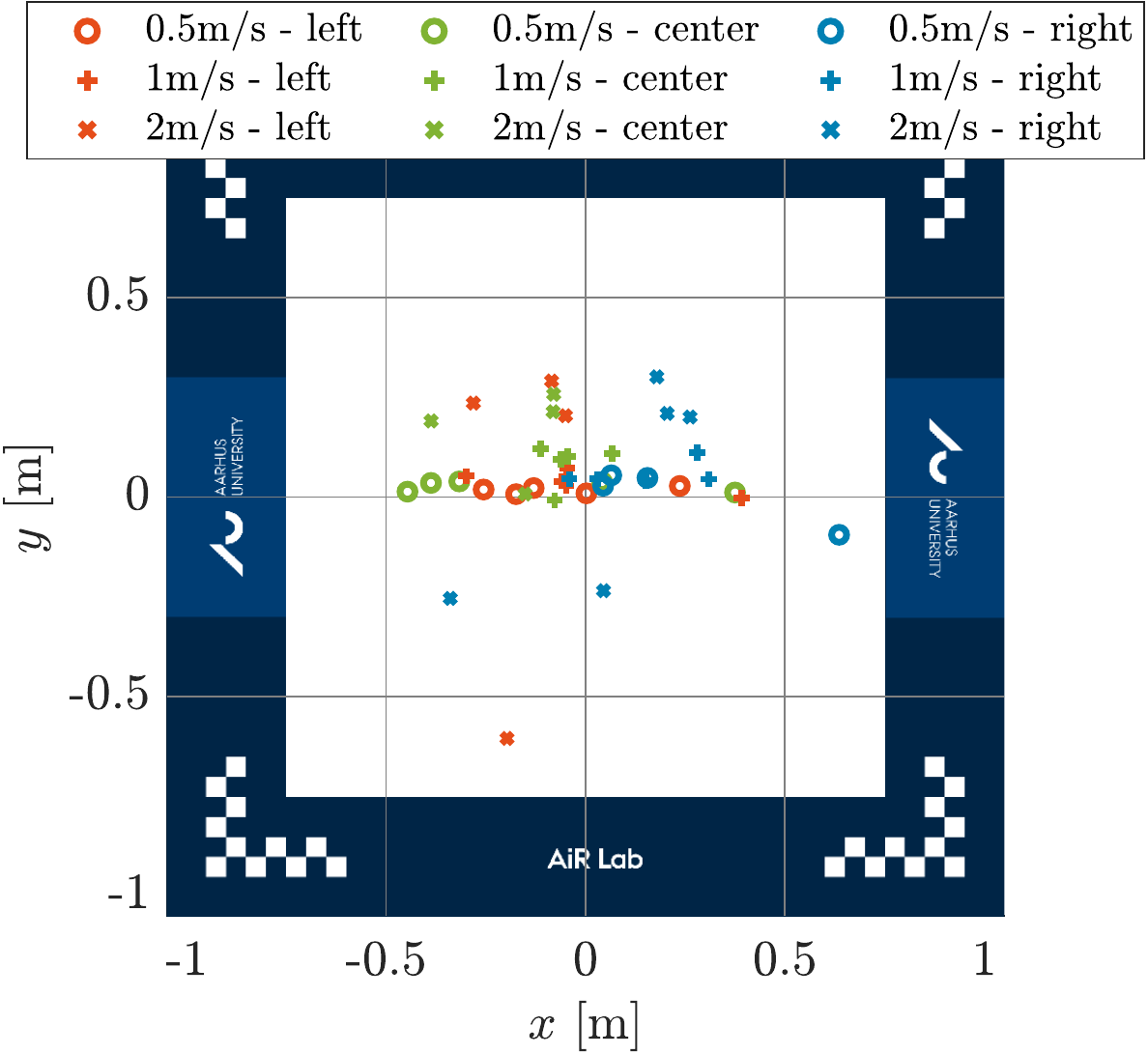}
\caption{Gate crossing points at different speeds from different starting
locations. For each speed-direction case, the experiments are repeated five
times. It can be seen that at lower speed the vertical error is lower; while at
higher speed the UAV tends to cross the gate in the upper part of the gate.}
\label{fig:gateCrossing_2D}
\end{figure}

For the statistical analysis of navigation performances, the
experiments are executed five times for
three different cruise speeds ($0.5 \si{m/s}$, $1 \si{m/s}$ and $2 \si{m/s}$) starting from three different locations (left, centre and right) for a total of $45$ runs.
A front view of the gate can be seen in Fig.~\ref{fig:gateCrossing_2D}, which
shows the points at which the drone has crossed the gate. It can be observed that flights at $0.5 \si{m/s}$ are
more stable and yield a more precise gate crossing than flights at $2
\si{m/s}$, where the points are more scattered and less clustered in the centre
of the gate.

\begin{figure}[!t]
\centering
\includegraphics[width=0.99\columnwidth]{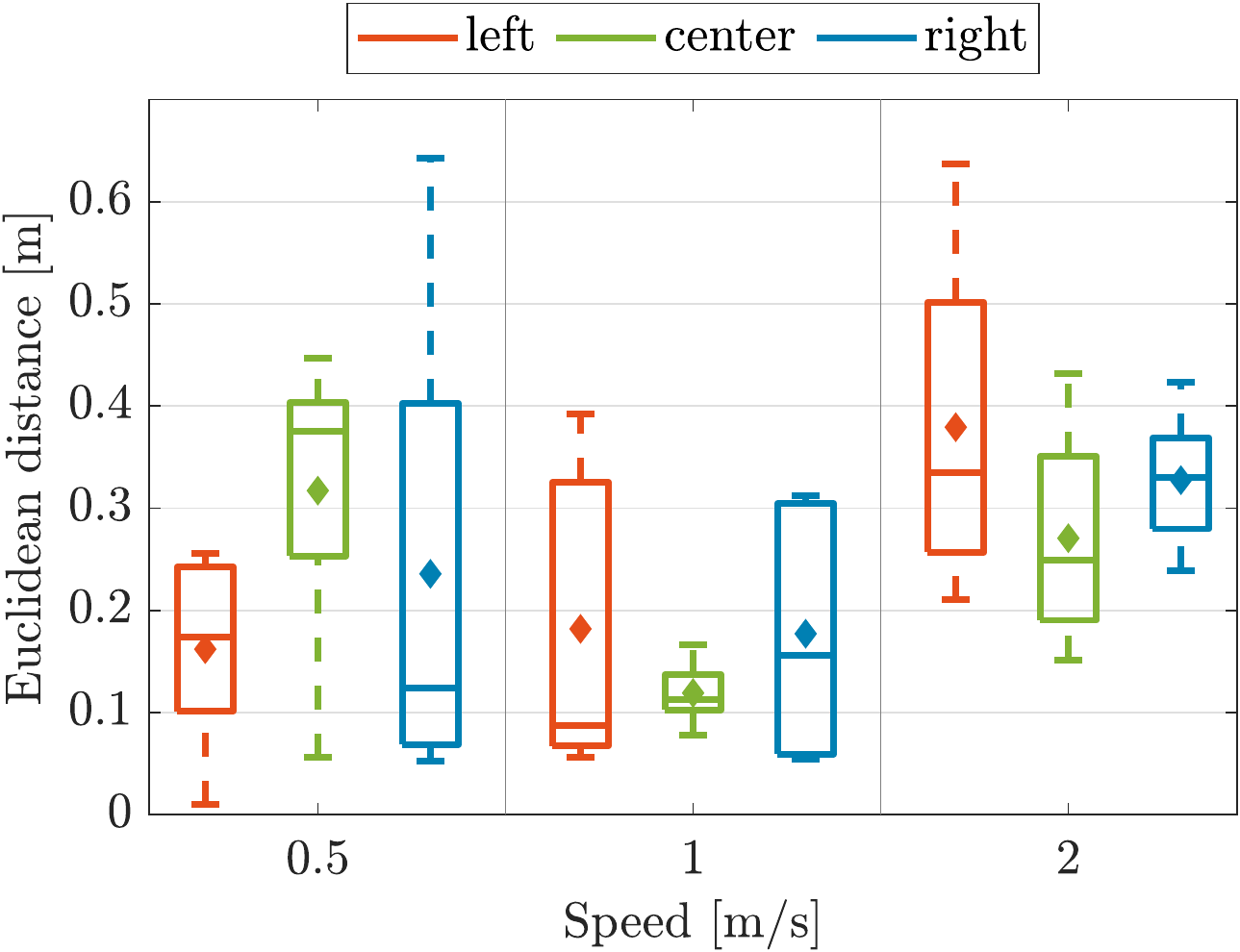}
\caption{Euclidean distance between the crossing point and the centre of the gate at different speeds from different initial position. The diamond represents the average Euclidean distance.}
\label{fig:gateCrossing_boxplot}
\end{figure}

The Euclidean distance between the crossing point and the centre of the gate
for different flying speed and starting locations is shown in
Fig.~\ref{fig:gateCrossing_boxplot}. It is seen that on average the deviation
from the centre of the gate for lower speeds ($0.5 \si{m/s}$ and $1 \si{m/s}$)
has comparable magnitude; while this quantity increases for faster flights ($2
\si{m/s}$) because of the coupled dynamics of the UAV. Besides, the flights
originating in front of the gate have predominantly lower fluctuation, since in
those cases the crossing gate area is larger.

\section{Conclusions and Future Work}
\label{sec:conclusion}

In this work, a hypothesis on the use of synthetic images to solve the gate
detection challenge in drone racing is proposed and put to the test.
A complete semi-synthetic dataset generation pipeline is
implemented, and this method's results are tested on a state-of-the-art
single-shot object detector. The model is evaluated on semi-synthetic images,
as well as pictures of real gates. The results show that there is no
significant domain shift impeding the precision of the detection. Furthermore,
a flight framework is designed, based on a PID controller
and a state machine. For the decision whether to cross the gate or keep
aligning with it, a second, smaller CNN, is designed and trained on hybrid
images to estimate the distance to the target object.
The system is tested in real conditions in unknown states,
and is able to successfully
cross the same gate at different speeds and initial points, with a high success
rate and a consistent trajectory. Thus, the proposed method for training
CNNs with synthetic imagery is viable and shows to be usable for high-speed UAV flight in unknown environments, while dramatically decreasing the time spent on the preparation of a dataset.

In the future, several improvements can be brought to the proposed framework, more specifically on the control side. Adding path planning to the controller would allow for a fluid flight
in a succession of gates. However, the orientation estimation of the gates must be implemented.

\section*{Supplementary Material}

The real-time experimental video is available at \href{https://youtu.be/T4gJgPNdiH8}{youtu.be/T4gJgPNdiH8}. The project's code, datasets and trained models are available at: \href{https://github.com/M4gicT0/autonomous-drone-racing}{github.com/M4gicT0/autonomous-drone-racing}.

\bibliographystyle{IEEEtran}
\bibliography{IEEEabrv,reference}

\end{document}